\documentclass{article}

\usepackage[preprint]{neurips_2023}

\usepackage[utf8]{inputenc} 
\usepackage[T1]{fontenc}    
\usepackage{hyperref}       
\usepackage{url}            
\usepackage{booktabs}       
\usepackage{amsfonts}       
\usepackage{nicefrac}       
\usepackage{microtype}      
\usepackage{xcolor}         

\usepackage{graphicx}
\usepackage{listings}
\usepackage{soul} 
\usepackage{array}
\usepackage{tabularx}
\usepackage{multirow}
\usepackage{mathtools}

\usepackage{graphicx}  
\usepackage{wrapfig}  

\usepackage{amsmath,amsthm, amssymb, amsfonts}

\usepackage{float}  
\usepackage{etoc}

\newcommand{\quotes}[1]{``#1''}

\lstnewenvironment{pythoncode}[1][]{
    \lstset{
        language=Python,
        basicstyle=\footnotesize\ttfamily,
        keywordstyle=\bfseries\color{blue},
        commentstyle=\color{gray},
        stringstyle=\color{red},
        showstringspaces=false,
        breaklines=true,
        frame=single,
        backgroundcolor=\color{white},
        captionpos=b,
        #1
    }
}{}

\theoremstyle{definition}

\theoremstyle{definition}

\title{IntelliGraphs: Datasets for Benchmarking \\ Knowledge Graph Generation}

\author{Thiviyan Thanapalasingam \\
University of Amsterdam \\
\texttt{thiviyan.t@gmail.com}
\And Emile van Krieken \\
Vrije Universiteit Amsterdam \\
\And Peter Bloem \\ Vrije Universiteit Amsterdam
\And Paul Groth \\ University of Amsterdam
}

\begin{document}

\maketitle

\begin{abstract}
Knowledge Graph Embedding (KGE) models are used to learn continuous representations of entities and relations. A key task in the literature is predicting missing links between entities. However, Knowledge Graphs are not just sets of links but also have semantics underlying their structure. Semantics is crucial in several downstream tasks, such as query answering or reasoning. 
We introduce the \emph{subgraph inference task}, where a model has to generate likely and semantically valid subgraphs. We propose \emph{IntelliGraphs}, a set of five new Knowledge Graph datasets. The IntelliGraphs datasets contain subgraphs with semantics expressed in logical rules for evaluating subgraph inference. We also present the dataset generator that produced the synthetic datasets. We designed four novel baseline models, which include three models based on traditional KGEs. We evaluate their expressiveness and show that these models cannot capture the semantics. We believe this benchmark will encourage the development of machine learning models that emphasize semantic understanding.
\end{abstract}

\section{Introduction}

Knowledge Graphs (KGs) contain knowledge about the world structured as graphs with entities connected through different relations \citep{hogan2021knowledge}. Large-scale KGs are widely used in a range of applications, such as query answering \citep{arakelyan2020complex} and information retrieval \citep{noy2019industry}.

To address the problem of incompleteness in KGs, Knowledge Graph Embedding (KGE) models were developed. These learn continuous representations for entities and relations \citep{bordes2013translating, yang2014embedding} through \emph{link prediction}, the task of predicting missing links in large KGs, by learning scoring functions that rank entities \citep{ruffinelli2019you}. These approaches implicitly assume that each link (also known as a \emph{triple}) in a Knowledge Graph can be predicted \emph{independently}. In this view of Knowledge Graphs, each triple is seen as a kind of ``atomic fact'' which is true or false independent of other triples.

\begin{figure}[h]
    \centering
    \includegraphics[width=0.9\linewidth]{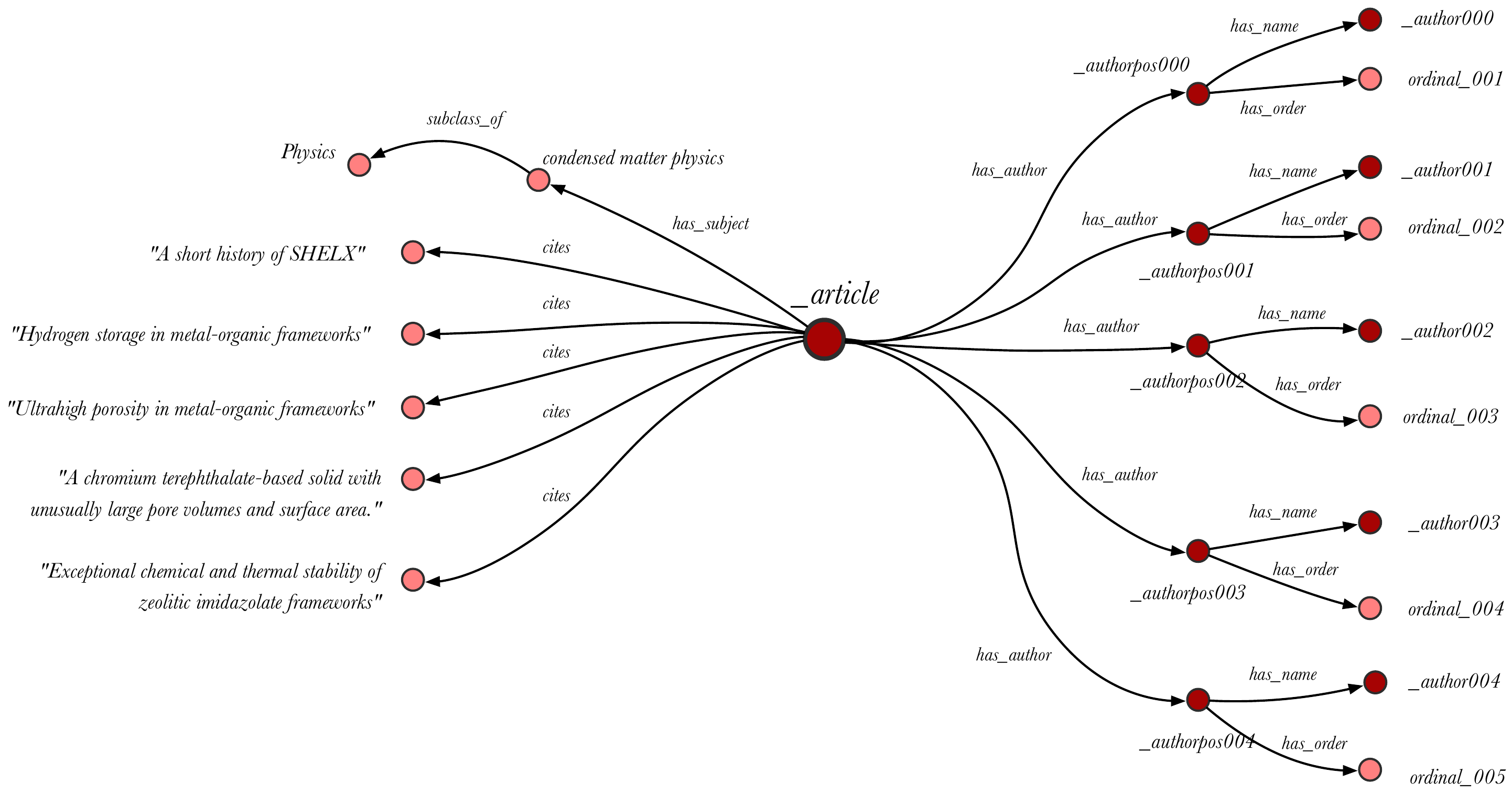}
    \caption{ A subgraph about a research article representing the complex interdepencies between the relevant entities. This example was extracted from Wikidata. The the lighter nodes represent entities and darker nodes represent existential nodes (explained in Section \ref{sec:intelligraphs}).}
    \label{fig:fg-example}
\end{figure}

However, in modern Knowledge Graphs, triples depend on each other. For example, the triples \texttt{value(temperature\_NY, 77)} and \texttt{unit(temperature\_NY, Fahrenheit)} together describe that the temperature in New York is 77 °F. In this case, the truth of the first triple depends on the content of the second. Figure~\ref{fig:fg-example} provides a more complex example which represents subgraph about a research article. In our work, we are interested in subgraphs that have many interdependent facts. 

Existing KGE models cannot capture such interdependencies between triples \citep{wen2016representation}. In this paper, we introduce a new task that can be used to evaluate models that generate several connected triples together, modelling their interdependencies. We call this task \emph{subgraph inference}. The idea is that where KGE models can be used to predict missing links, subgraph inference models could be used to predict missing subgraphs: sets of interdependent links.

To simplify the problem of subgraph inference, we assume that a set of true subgraphs is provided so that the problem reduces to training a generative model on small knowledge graphs over a shared set of entities and relations.\footnotemark Such predictions must not only capture the general structure of the graph, but they must also allow us to generalize effectively to graphs not explicitly shown in the data. 

\footnotetext{That is, for a true generalization of link prediction to subgraph prediction, we would provide a single, large knowledge graph and require a model to predict missing subgraphs. In the interest of separating concerns, we ask here only if generative models over small knowledge graphs are feasible.}

So far, the lack of datasets with well-understood semantics has hampered studying how effectively KGE models capture semantics. Existing datasets commonly used for benchmarking KGE models, such as FB15k-237 and WN18RR, lack sufficient logical constraints to investigate semantics thoroughly. Logical constraints play an important role as they help maintain the logical consistency of facts in a structured knowledge base. 

Thus, the three main contributions of our work are as follows:

\begin{enumerate}
    \item \textbf{Subgraph Inference.} We define a new task, where the goal is to generate, from a set of examples, novel subgraphs that follow certain logical rules. We specified new evaluation metrics that help empirically assess generated graphs' semantic validity and novelty.
    \item \textbf{IntelliGraphs}
    \begin{enumerate}
        \item \textbf{Synthetic Datasets.} We propose three synthetic datasets, each designed to capture different levels of semantics. We also describe the underlying semantics using First-Order Logic.
        \item \textbf{Real-world Datasets.} We extract subgraphs from Wikidata according to simple basic patterns to generate two \emph{real-world} datasets.\footnote{https://www.wikidata.org/}
    \end{enumerate}
    \item \textbf{Data Generator.} We developed a Python package that randomly generates and verifies subgraphs using pre-defined logical constraints.
\end{enumerate}
The datasets and generators are publicly available on: \url{https://github.com/thiviyanT/IntelliGraphs}. The generator is available as a Python package which can be installed through PyPy, and Conda package managers. \footnote{ PyPy: \url{https://pypi.org/project/intelligraphs} \& Conda:\url{https://anaconda.org/thiv/intelligraphs} } To ensure long-term preservation and easy access, we made the datasets available on Zenodo. \footnote{ \url{https://doi.org/10.5281/zenodo.7824818} } 

\section{Benchmark Tasks}

In this Section, we discuss related benchmark tasks and their limitations, and then, we introduce a novel research task we call \emph{subgraph inference}.

\subsection{Limitations of Link Predictors}
\label{section:lplimitations}

\textbf{Binary relations} \quad KGE models exploit structural regularities to perform Knowledge Graph completion. The last decade has seen the developments of several KGE models \citep{ruffinelli2019you}, which predict the likelihood that a pair of entities are related by a given binary relation. However, a set of binary relations cannot represent an N-\emph{ary} relation because the links depend on each other. Regardless of the context, KGE models assign a set of probabilities on links, and those probabilities are independent of each other.

\textbf{N-\emph{ary} relations} \quad Link prediction has been extended to cover N-\emph{ary} relations, where the goal is to predict a missing link in an N-\emph{ary} fact. N-\emph{ary} relation can operate on any arbitrary number of entities. Modelling N-\emph{ary} relations as triples and treating them as entities in binary relations results in a loss of
structural information \citep{wen2016representation}.  \citet{wen2016representation} define N-\emph{ary} relations as the mappings from the attribute sequences to the attribute values, such that each N-\emph{ary} fact is an instance of the corresponding N-\emph{ary} relation. GRAN is a graph-based approach which uses a Transformer decoder to score N-\emph{ary} facts \citep{wang-etal-2021-link}. NeuInfer uses fully-connected neural networks to embed N-\emph{ary} relations and score candidate triples \citep{guan-etal-2020-neuinfer}. These models were evaluated by inferring an element in an N-\emph{ary} fact. Because a single N-ary relation can be represented in a set of binary relations (i.e. triples), subgraphs can be used to represent N-ary relations. This means that Subgraph models could be used to solve N-ary relation prediction, but the task is strictly broader than that: every single N-ary relation can be represented as a subgraph, but not every class of subgraphs can be naturally captured by a single n-ary relation.

\textbf{Link prediction evaluation} \quad 
The standard link prediction evaluation framework \citep{ruffinelli2019you} uses ranking-based evaluation metrics, such as Hits@k and Mean Reciprocal Rank, which do not explicitly check for the semantics of the predicted links. Instead, the evaluation protocol assumes that the underlying semantics can be indirectly validated if a missing link has been correctly predicted.
In our work, we set out to \emph{explicitly} check the semantics of newly generated subgraphs.

\subsection{Subgraph Inference}

A \emph{Knowledge Graph}, $G$, is a tuple $G = (V,E, \mathcal{E}, \mathcal{R}, L)$. $E$ is a set of edges where $E = V \times \mathcal{R} \times V$ and $\mathcal{R}$ is the set of relations. $V$ is set of nodes drawn from the set of possible entities $\mathcal{E}$ in $G$.  $L$ is a set of functions that define the semantics of $G$ by determining which structures are permissible or not in $G$.

Given a Knowledge Graph $G$,  we call a \emph{subgraph} $F$ a tuple $\left(V^{f}, E^{f}, \mathcal{R}\right)$ where $V^{f} \subset V$ and $E^{f}=\left\{ (u, r, v) \mid u \in V^{f}, v \in V^{f}\right\}$, $r \in \mathcal{R}$ and $(u, r, v) \in E$. We require subgraphs to be connected graphs. Every subgraph complies with the semantics of the Knowledge Graph, $L_{G}$. 

For example, using the example from the introduction, the statement \quotes{The temperature is 77°F.}  can be expressed as a simple two-triple subgraph $F$ of some larger graph with the triples \texttt{value(temperature, 77)}, \texttt{unit(temperature, Fahrenheit)}.\footnotemark The meaning of the entity \texttt{temperature} is dependent on both triples. Notably, the unit \quotes{Fahrenheit} gives value \quotes{77} additional context. Here, the semantics could be captured by a function that checks that $\texttt{is\_instance(value, integer)} \wedge \texttt{is\_unit(unit, units\_of\_temperature)}$.

\footnotetext{We do not specify what larger graph this graph is a subgraph \emph{of}. In most of our tasks, only the set of entities and relations of the larger graph is given, and the rest of the graph is left implicit.}


\textbf{Problem Statement} \quad \emph{Subgraph Inference} is the task of inferring  missing subgraphs given a set of existing subgraphs from a Knowledge Graph, $G$.\footnotemark The inferred subgraphs must adhere to the semantics of the original KG. We define the task as follows: Given a set of known subgraphs $S_{G}^{k}$ from a given Knowledge Graph $G$, infer missing subgraphs $S_{G}^{m}$ that comply with the same logical constraints of the initial Knowledge Graph $L_{G}$. We assume we have access to $L_G$. The model does not have access to $L_G$ during learning. Indstead, it is used to evaluate the semantics of the model output during evaluation. 

\footnotetext{In symbolic AI, the term \emph{inference} refers to \emph{formal reasoning}. Here we use it to mean estimating the probability distribution over the model's unobserved variables given observed data (\emph{i.e.} probabilistic inference). }

These subgraphs can be added back to the KG; therefore, this task can be seen as an extension of link prediction. To make this extension complete, we should also specify how the training subgraphs are extracted from $G$. However, to isolate the question of generative modelling of knowledge graphs, we take this process as given in our tasks. For instance, in the two real-world datasets, we extract subgraphs from Wikidata according to a hand-designed pattern. In the synthetic datasets, we simply provide a set of small knowledge graphs over a shared set of entities and relations, leaving the larger graph $G$ entirely implicit. With this choice, the task reduces to training a generative model over small knowledge graphs with a shared set of entities and relations.

\textbf{Key Challenge.} \quad The subgraphs need to adhere to specific semantics which, in a learning setting, have to be inferred from a limited set of examples, such as learning the types of entities.

\section{IntelliGraphs}
\label{sec:intelligraphs}

Motivated by the aforementioned limitations of link prediction datasets and the new task of subgraph inference, we introduce five new benchmark datasets where each dataset tests different semantics. Table \ref{dataset-stats} shows key statistics about the synthetic and real-world graphs. The appendix (see Section \ref{subsec:benchmarks}) describes the algorithm used to generate the datasets. 

\textbf{Data Generator} \quad The sampler $D$ samples subgraphs according to a probability distribution $P$ defined in the Python implementation of IntelliGraphs. For each dataset's logical constraints $L$, the sampler samples a graph $F$ from the probability distribution $P$, ensuring that $F$ satisfies all the logical constraints in $L$. 

\textbf{Existential Nodes} \quad In some settings, it is necessary to have nodes that refer to entities that only occur in one instance. For example, in the \texttt{wd-movies} dataset introduced below, each subgraph in the data represents a movie. Its actors, directors and genres are entities that occur in multiple instances, so a model can learn representations by observing the different contexts in which they occur. However, each instance also contains one node representing the movie the graph describes. These only occur in one instance, so a model cannot learn a representation for the specific instance, only a general representation which expresses that some movie exists for which this subgraph is true. We call such nodes \emph{existential nodes} (in analogy to existentially quantified variables in logical formulas) and use a special label, such as \texttt{\_movie}, to refer to them in all instances.\footnotemark Strictly speaking, this turns the predicted subgraphs into subgraph \emph{patterns} of the Knowledge Graph $G$, but we refer to them as subgraphs to keep the terminology simple.

\footnotetext{For most models, the difference will only be in the interpretation. For example, our baseline models will learn one embedding vector for the node labelled \texttt{\_movie}, which we use wherever movies occur. As such, we do not treat it differently from the node labelled \texttt{Antonio\_Banderas}, although when we interpret the graph, these nodes mean different things.}

\subsection{Semantics}

We use First-Order Logic (FOL) to express the underlying logical rules of the datasets. These logical rules, $L$, were hand-crafted for subgraphs for every dataset, and we ensured that all subgraphs complied with the logical rules. Section \ref{appx-semantics} (in the Appendix) provides a complete set of logical constraints for each IntelliGraphs dataset.

\noindent \textbf{Logical Constraint Verifier.} The \emph{Logical Constraint Verifier} $v$ is a function that verifies whether the logical constraints $L$ hold in a generated subgraph $F$. We wrote a logic constraint verifier within the IntelliGraphs Python package.\footnotemark The logical constraint verifier $v(F, L)$ returns true if and only if the subgraph $F$ is consistent with all logical rules $L$.

\footnotetext{A reasoning engine could also be used for checking the subgraphs for logical consistency. We wrote a set of functions in Python for constraint verification and embedded it into the IntelliGraphs Python package to easily verify graphs without loading them into a reasoning engine.}

\begin{table}[t]
\caption{The size of the training, validation and test split for the five datasets used in this work. The number of edges is fixed for the synthetic datasets and is variable for the Wikidata-based graphs.}
\label{dataset-stats}
\begin{center}
\begin{tabular}{lcccc}
\hline
\multicolumn{1}{c}{\bf Dataset}  &\multicolumn{1}{c}{\bf Split  (train/val/test)}&\multicolumn{1}{c}{\bf Entities}&\multicolumn{1}{c}{\bf Relations}&\multicolumn{1}{c}{\bf Triples }\\ 
\hline
syn-paths & 60000/20000/20000 & 49 & 3 & 3\\
syn-types & 60000/20000/20000 & 30 & 3 & 3\\
syn-tipr & 50000/10000/10000 & 130 & 5 & 5\\
wd-movies & 38267/15698/15796 & 24093 & 3 & 2 -- 21\\
wd-articles & 54163/22922/22915 & 60932 & 6 & 4 -- 212\\
\hline
\end{tabular}
\end{center}
\end{table}

\subsection{Synthetic Datasets}

Synthetic datasets allow complete control over the problem setup and provide a convenient testbed for developing new machine learning models. The dataset is generated by the generator, $D$. We checked if the generated subgraphs satisfy the logical rules $L$.\footnotemark Here is a brief description of the synthetic datasets:

\footnotetext{It is important to note that logical consistency does not equate to factual accuracy. We simply want to ensure that the generated dataset is consistent with the logical rules.}

\begin{itemize}
    \item \texttt{syn-paths} is a dataset with path graphs. Path graphs have simple semantics that can be algorithmically verified in linear time. Path graphs have a single directed path of length 3 and no other edges.

    \item \texttt{syn-types} contains entities with types \texttt{Language}, \texttt{Country} and \texttt{City}. These are connected by three relations according to the relation's type constraints: $\verb|sam_type_as|$ can only exist between the same entity types, $\verb|could_be_part_of|$ between a capital city and country, and $\verb|could_be_spoken_in|$ between a language and a country. The connections are otherwise random.

    \item \texttt{syn-tipr} contains subgraphs based on the \emph{Time-indexed Person Role} (tipr) ontology pattern.\footnote{ \url{http://ontologydesignpatterns.org/wiki/Submissions:Time\_indexed\_person\_role} }  Here, the semantics are defined by the tipr graph pattern. The semantics include the fact that the start of an interval must precede its end.
    
\end{itemize}

\subsection{Real-World Datasets}

Wikidata \citep{vrandevcic2014wikidata} is a large graph-structured knowledge base which consists of crowdsourced factual knowledge on various topics.\footnote{ \url{https://www.wikidata.org} } We created two datasets from Wikidata using specific graph patterns to extract subgraphs about movies and research articles. Here is a brief description of the two datasets:

\begin{itemize}
    \item \texttt{wd-movies} contains small graphs extracted from Wikidata that describe movies. Each graph contains one existential node representing the movie, entity nodes for the movie's director(s) connected by a \texttt{has\_director} relation, entity nodes for the movie's cast connected by a \texttt{has\_actor} relation and an entity for the movie's genre connected by a \texttt{has\_genre} relation. 

    \item \texttt{wd-articles} contains small graphs that describe research articles extracted from Wikidata. Each article is annotated by an ordered list of authors, implemented by a blank node for each author linked to a node representing the author and to a node representing the order in the author list. We add a list of the other articles that the current article references, and a list of subjects, together with selected superclasses of those subjects. In this dataset, most node types, including the article's node, may be existential or entity nodes.
\end{itemize}

\section{Evaluation}

\subsection{Evaluation by bits-per-graph}

The most common objective for a generative model is probably maximum likelihood: the probability of a graph from the test data under the model should have maximal probability, or, equivalently, minimal negative log probability. When base $2$ logarithms are used, the latter quantity, $-\log_2 p(S, E)$, can be interpreted as the number of bits required to compress the graph \citep{rissanen1978modeling, grunwald2007minimum}. Averaging over all graphs, we arrive at a metric of \emph{bits-per-graph} to evaluate how well our model satisfies the maximum likelihood objective. 


\subsection{Semantics}

We evaluate the semantics of graphs generated by our baseline models using the following evaluation metrics: 1)  \emph{\% Valid Graphs} is the probability of sampling graphs that are logically valid according to the logical constraints for each dataset, 2) \emph{\% Novel Graphs} is the probability of sampling graphs that are not in the training data, 3) \emph{\% Novel \& Valid Graphs} is the probability of sampling graphs that are logically valid and are not in the training data, and 4) \emph{\% Empty Graphs} is the probability of sampling graphs that did not yield any graphs, due to either $p(E)$ or $p(S\mid E)$ being too low. An ideal model gives a high probability of sampling logically valid graphs but uses a minimal number of code lengths to compress graphs. 

\subsection{Baseline Models}

To the best of our knowledge, no probabilistic models in the literature can infer new subgraphs for knowledge graphs. Therefore, we developed a set of simple baselines inspired by traditional KGE models: ComplEx \citep{trouillon2016complex}, DistMult \citep{yang2014embedding} and TransE \citep{bordes2013translating}. Traditional KGE models are trained to rank all possible triples to give the correct triple the highest score \citep{ruffinelli2019you}. ComplEx, DistMult and TransE all use different scoring functions. TransE represents relations as translation between entities, whereas DistMult models relations as bilinear interactions. ComplEx extends DistMult using complex-valued embeddings. 

We model a subgraph $F$ by decomposing it into its entities and structure $F=(E, S)$, that is, $p(F) =  p(S\mid E) \;p(E)$. Unlike traditional KGE models, we train our baseline models with a maximum likelihood objective.

We decompose the objective function as follows:
\begin{equation}
    - \log_2 p(F)= -\log_2 p(S|E) - \log_2 p(E). 
    \label{eq:two-part}
\end{equation}

Each of the terms in Equation~\ref{eq:two-part} can be read as separate codelengths: $-\log_2 P(E)$ describes the bits required to encode the entities, and $-\log_2 P(S\mid E)$ describes the bits required to encode the structure once the entities are known.

We model $p(E)=\prod_{e\in E} p(e)$, with $p(e)$ estimated as the relative frequency of $e$ in the training data (the proportion of training subgraphs it occurs in). We train KGE models to estimate $p(S \mid E)$. We use 
\begin{equation}
p(S \mid E) = \prod_{(s, p, o) \in S_T} p((s, p, o) \mid E ) \prod_{(s, p, o)\in S_N} 1 - p((s, p, o) \mid E)\text{,}
\end{equation}
where $S_T$ represents the triples in the subgraph $F$, and $S_N$ represents all possible triples that are not in the subgraph (\emph{i.e.} all possible \emph{negatives}).

Our \emph{random} baseline model generates a random graph prediction by sampling  $p(E)$ and $p(S|E)$ from a uniform distribution. It then computes the exact number of bits required to represent these probabilities, using $-log_2(p)$ to determine the entropy of each probability value. This model does not need to be trained.

Table \ref{graph-compression-results} shows that the KGE baselines learn more compact representations than the random baseline. The ComplEx baseline is most effective at compressing the structure of these graphs $p(S|E)$, despite requiring real and imaginary parts. The scale of complexity, represented by code length, seems to increase rapidly from synthetic to real-world datasets. For instance, the highest code length for \texttt{syn-paths} is $69.51$ (for the TransE baseline), while the lowest code length for \texttt{wd-movies} is $202.68$ (for ComplEx). \texttt{wd-movies} and \texttt{wd-articles} have many more entities to sample, making them more challenging to compress.

\begin{table}
  \caption{Estimate of the codelengths, $-\log_2 p(F)$, (the number of bits) required to compress a graph using the four baseline models for IntelliGraphs datasets. We used the test split for this. We rounded the numbers up/down to two decimal points. }
  \label{graph-compression-results}
  \centering
    \begin{tabular}{c c c c c}
    \toprule
        Datasets & Baseline Models & $-\log_2 p(S|E)$ &  $-\log_2 p(E)$ & $C(S,E)$ \\ \hline 
        & \text{\emph{random}} & 95.94 & 98.04 & 193.98 \\ 
        \text{\texttt{syn-paths}} & \text{TransE} & 16.19 & 33.69
 & 49.89 \\ 
        &  \text{DistMult} & 14.90
 & 33.69 & 48.58 \\ 
        &  \text{ComplEx} & 20.71 & 33.69 & 54.39 \\ 
        \hline 
        & \text{\emph{random}} & 360.27 & 259.80 & 620.08 \\ 
        \text{\texttt{syn-tipr}} & \text{TransE} & 28.70 & 40.81 & 69.51
 \\ 
        & \text{DistMult} & 26.70 & 40.81 & 67.51 \\ 
        & \text{ComplEx} & 23.15 & 40.81 & 63.96 \\ 
        \hline 
        & \text{\emph{random}} & 187.11 & 59.99 & 247.1 \\ 
        \text{\texttt{syn-types}} & \text{TransE} & 19.05 & 29.21 & 48.26 \\ 
        & \text{DistMult} & 18.24 & 29.21 & 47.46 \\ 
        & \text{ComplEx} & 18.48 & 29.21 & 47.69 \\ 
        \hline 
        & \text{\emph{random}} & 483.81 & 48185.97 & 48669.78 \\
        \text{\texttt{wd-movies}}  & \text{TransE} & 51.39 & 157.21 & 208.60 \\ 
        & \text{DistMult} & 51.29 & 157.21 & 208.50 \\ 
        & \text{ComplEx} & 45.46 & 157.21 & 202.68\\ 
        \hline
        & \text{\emph{random}} & 12623.20 & 122366.51 & 134989.71 \\
        \text{\texttt{wd-articles}} & \text{TransE} & 280.67 & 629.98 & 910.65 \\ 
        & \text{DistMult} & 271.94 & 629.98 & 901.91 \\ 
        & \text{ComplEx} & 257.33 & 629.98 & 887.30 \\ 
    \bottomrule
    \end {tabular}
\end{table}

\subsection{Subgraph Inference}

Table \ref{entities-sampling-results} shows the probabilities of sampling graphs that are logically consistent. We perform subgraph inference under two different settings: 

\nopagebreak

\begin{itemize}
    \item \textbf{Sampling $P(E)$ and $P(S\mid E)$.} Here, the baseline models sample both the entities that are relevant for a subgraph and infer their edge connectivity. Our results indicate that the probability of sampling valid graphs is consistently $0\%$. Selecting the incorrect entities negatively impacts the structure prediction. Our results indicate that this task is challenging, especially for the random baseline, as it consistently fails to infer valid subgraphs.
    
    \item \textbf{Sampling only $P(S\mid E)$.} In this setup, the model is given an advantage by having access to the correct set of entities (\emph{i.e.}, we give $p(E)$), such that it only needs to predict the edge connections between the given entities. It is worth noting that under this setting, the baseline model collapses into a link predictor as it just predicts the edge connections between the given entities. Despite giving the advantage, the baseline models could not generate many logically consistent subgraphs. Interestingly, this also reveals the complexity of the datasets and what semantics these KGE models can learn. Most KGE models are able to generate some valid path graphs, while for \texttt{syn-tipr}, which requires some temporal reasoning, seems more challenging for all baseline models. Inferring the correct entity types from \texttt{syn-types} was possible for a few graphs.
\end{itemize}

\begin{table}[t]
\caption{ Semantic validity of the graphs produced by our baseline models. High values for \emph{\% Novel \& Valid Graphs} is desirable. We have tested subgraph inference under two settings: 1) Sampling from \emph{both} $P(E)$ and $P(S\mid E)$, and 2) Sampling from $P(S\mid E)$ \emph{only}, taking $E$ from the test data. We check the novelty of the sampled graphs by comparing them against the training and validation set. We used the same hyperparameters from the model compression experiments here. Best performing models for each dataset is \textbf{bolded}.}
\label{entities-sampling-results}
\begin{center}
\begin{tabular}{cllcccc}
\hline
& & & \% & \% Novel & \% & \% \\
Setting & Dataset & Model & Valid & \& Valid & Novel & Empty \\
& & & Graphs & Graphs & Graphs & Graphs \\
\hline
\multirow{20}{*}{\rotatebox[origin=c]{90}{ {\large Sampling from $P(E)$ and $P(S \mid E)$}}}
& \multirow{4}{*}{\texttt{syn-paths}}
& random & 0 & 0 & 100 & 0 \\
& & TransE & 0.25 & 0.25 & 23.45 & 76.55 \\
& & DistMult & 0.69 & 0.69 & 14.59 & 85.41 \\
& & \textbf{ComplEx} & \textbf{0.71} & \textbf{0.71} & \textbf{14.27} & \textbf{85.73} \\
\cline{3-7}
& \multirow{4}{*}{\texttt{syn-tipr}}
& random & 0 & 0 & 100 & 0 \\
& & TransE & 0 & 0 & 5.58 & 94.42 \\
& & DistMult & 0 & 0 & 13.34 & 86.66 \\
& & ComplEx & 0 & 0 & 4.95 & 96.05 \\
\cline{3-7}
& \multirow{4}{*}{\texttt{syn-types}}
& random & 0 & 0 & 100 & 0 \\
& & \textbf{TransE} & \textbf{0.21} & \textbf{0.21} & \textbf{15.44} & \textbf{84.56} \\
& & DistMult & 0.13 & 0.13 & 12.46 & 87.53 \\
& & ComplEx & 0.07 & 0.07 & 10.25 & 89.75 \\
\cline{3-7}
& \multirow{4}{*}{\texttt{wd-movies}}
& random & 0 & 0 & 100 & 0 \\
& & TransE & 0 & 0 & 14.61 & 85.39 \\
& & DistMult & 0 & 0 & 12.93 & 87.07 \\
& & ComplEx & 0 & 0 & 1.87 & 98.13 \\
\cline{3-7}
& \multirow{4}{*}{\texttt{wd-articles}}
& random & 0 & 0 & 100 & 0 \\
& & TransE & 0 & 0 & 4.58 & 95.42 \\
& & DistMult & 0 & 0 & 0 & 100.00 \\
& & ComplEx & 0 & 0 & 2.46 & 97.54 \\
\hline
\multirow{20}{*}{\rotatebox[origin=c]{90}{ {\large Sampling from $P(S \mid E)$ only}}}
& \multirow{4}{*}{\texttt{syn-paths}}
& random & 0 & 0 & 100 & 0 \\
& & TransE & 5.25 & 5.25 & 95.52 & 4.48 \\
& & DistMult & 9.69 & 9.69 & 95.28 & 4.71 \\
& & \textbf{ComplEx} & \textbf{10.10} & \textbf{10.10} & \textbf{95.58} & \textbf{4.42} \\
\cline{3-7}
& \multirow{4}{*}{\texttt{syn-tipr}}
& random & 0 & 0 & 100 & 0 \\
& & TransE & 0 & 0 & 99.45 & 0.55 \\
& & DistMult & 0 & 0 & 99.43 & 0.57 \\
& & ComplEx & 0 & 0 & 99.64 & 0.36 \\
\cline{3-7}
& \multirow{4}{*}{\texttt{syn-types}}
& random & 0 & 0 & 100 & 0 \\
& & TransE & 1.43 & 1.43 & 95.42 & 4.58 \\
& & \textbf{DistMult} & \textbf{1.44} & \textbf{1.44} & \textbf{96.19} & \textbf{4.81}  \\
& & ComplEx & 1.01 & 1.01 & 94.17 & 5.83 \\
\cline{3-7}
& \multirow{4}{*}{\texttt{wd-movies}}
& random & 0 & 0 & 100 & 0 \\
& & TransE & 0.07 & 0.07 & 97.01 & 2.99 \\
& & DistMult & 0.10 & 0.10 & 95.86 & 4.17 \\
& & \textbf{ComplEx} & \textbf{0.41} & \textbf{0.41} & \textbf{93.04} & \textbf{6.96} \\
\cline{3-7}
& \multirow{4}{*}{\texttt{wd-articles}}
& random & 0 & 0 & 100 & 0 \\
& & TransE & 0 & 0 & 98.35 & 1.65 \\
& & DistMult & 0 & 0 & 98.77 & 1.23 \\
& & ComplEx & 0 & 0 & 100.00 & 0.00 \\
\hline
\end{tabular}
\end{center}
\end{table}

\subsection{N-ary Link Prediction}

The simplest tasks in IntelliGraphs can indeed be modeled directly as an N-ary link prediction problem. For instance, the \texttt{syn-paths} graphs is a 4-ary hypergraph, and predicting hyperlinks on this graph is one way to solve the problem. However, as the tasks increase in complexity, the limitations of N-ary link prediction become clear. To model a task like \texttt{wd-articles} purely with link prediction, a single n-ary relation would need to capture the entirety of one subgraph describing an article, its author in order, the subjects of the article, and other features of the subgraph. Moreover, for the more complex tasks, the size of the subgraph is variable, which means that a single N-ary relation cannot capture the entire subgraph, unless the arity of the relation is somehow made variable. We leave the empricial study for future work. 

\section{Related Work}
\label{related-work-section}

\noindent \textbf{Datasets for Query Embedding.} Query Embedding (QE) involves interpreting complex logical queries, commonly represented as a small graph, and evaluated on QE datasets, such as GQE \citep{hamilton2018embedding}, Query2Box \citep{ren2020query2box}, and BetaE \citep{ren2020beta}. \citet{ren2023neural} presents a comprehensive comparison of datasets. As \citet{ren2023neural} highlight in their recent surrvey, query embedding datasets lack logical rules and types. Although the datasets in IntelliGraphs are similar to query embedding datasets, there is a difference in the purpose and applications. Our datasets can be used for learning distributions to infer new logically consistent subgraphs. In contrast, QA datasets are concerned with reasoning using logical rules to find a missing entity.

\noindent \textbf{Datasets for n-\emph{ary} Relations.} N-ary relations are relations involving more than two entities. Various methods have been studied in the literature that embeds complex N-ary relations, often in non-euclidean spaces \citep{wang-etal-2021-link, wen2016representation}. The difference between N-\emph{ary} relations and subgraphs is explained in Section~\ref{section:lplimitations}.

\noindent \textbf{Datasets for Neurosymbolic methods.} If we interpret knowledge graphs as a set of logical statements, we can see that the task of subgraph prediction is a neurosymbolic method: it combines symbolic systems with neural networks. Datasets have been proposed to test various aspects of such systems: interpretability, reasoning, and generalization capabilities. Several datasets were proposed to evaluate the understanding and reasoning of complex rules and abstract concepts. Table \ref{tab:comparison} (in the appendix) compares different datasets for Neurosymbolic AI from the literature. Existing datasets focus primarily on the image and text modalities, neglecting background knowledge expressed in graphs. 


\section{Conclusion}

Existing KG datasets used for representation learning lack well-understood semantics, which limits studying how well KGE models capture new semantics. In our work, we propose \emph{Subgraph Inference} as a new research problem and \emph{IntelliGraphs}, a collection of five new datasets for benchmarking models. Furthermore, we used baseline models inspired by traditional KGE models to estimate the code lengths of these graphs and sample logically valid subgraphs. Our findings show that traditional KGE models show a limited understanding of semantics after training. We observed a rapid increase in complexity, represented by code lengths, from synthetic to real-world datasets. This complexity makes real-world datasets more challenging to compress, which is an essential consideration for future research in graph compression. We found that the probability of sampling valid graphs was consistently low, emphasizing the complexity and difficulty of the task.

\textbf{Limitations.} \emph{Subgraph inference} assumes that the semantics of a KG is known. However, in some cases, this assumption may not hold. Furthermore, our datasets assume we test the machine learning models in a transductive setting; entities and relations not seen during training will not be handled well.

\textbf{Ethics Statement.} Our synthetic graphs are based on the logical rules we constructed and should not be used for applications where factual accuracy matters. However, \texttt{wd-movies} and \texttt{wd-articles} are based on real-world factual knowledge retrieved from Wikidata. Therefore, certain biases may be inherited from Wikidata. Since these datasets are likely unsuitable for training production models or for pretraining, we do not expect that these biases will ever affect systems making real-world decisions. Transparency about dataset creation and maintenance is critical for adopting new machine learning datasets \citep{gebru2021datasheets}. In the appendix, we provide a data card for IntelliGraphs to provide further information about the datasets.

\textbf{Applications of IntelliGraphs.} It is imperative to have guarantees for safety-critical applications to prevent machine learning models from making fatal mistakes. To develop these systems, datasets with logical constraints are helpful. In some problem domains, there is little or no data available such as cases where training machine learning models on sensitive data for medical or industrial use cases. While we do not provide datasets that are directly applicable to these use cases, IntelliGraph's dataset generation framework can be used to generate synthetic datasets using background knowledge about the problem domain. 


\paragraph{Acknowledgements} We would like to thank Frank van Harmelen and Patrick Koopmann for their feedback on this work. 

\bibliographystyle{plainnat} 
\bibliography{refs1} 

\begin{thebibliography}{32}
\providecommand{\natexlab}[1]{#1}
\providecommand{\url}[1]{\texttt{#1}}
\expandafter\ifx\csname urlstyle\endcsname\relax
  \providecommand{\doi}[1]{doi: #1}\else
  \providecommand{\doi}{doi: \begingroup \urlstyle{rm}\Url}\fi

\bibitem[Arakelyan et~al.(2020)Arakelyan, Daza, Minervini, and
  Cochez]{arakelyan2020complex}
Erik Arakelyan, Daniel Daza, Pasquale Minervini, and Michael Cochez.
\newblock Complex query answering with neural link predictors.
\newblock In \emph{International Conference on Learning Representations
  (ICLR)}, 2020.

\bibitem[Bordes et~al.(2013)Bordes, Usunier, Garcia-Duran, Weston, and
  Yakhnenko]{bordes2013translating}
Antoine Bordes, Nicolas Usunier, Alberto Garcia-Duran, Jason Weston, and Oksana
  Yakhnenko.
\newblock Translating embeddings for modeling multi-relational data.
\newblock In \emph{Neural Information Processing Systems (NIPS)}, pages 1--9,
  2013.

\bibitem[Bowman et~al.(2015)Bowman, Angeli, Potts, and
  Manning]{bowman2015large}
Samuel~R Bowman, Gabor Angeli, Christopher Potts, and Christopher~D Manning.
\newblock A large annotated corpus for learning natural language inference.
\newblock In \emph{Proceedings of the 2015 Conference on Empirical Methods in
  Natural Language Processing (EMNLP)}, pages 632--642, 2015.

\bibitem[Clark et~al.(2018)Clark, Cowhey, Etzioni, Khot, Sabharwal, Schoenick,
  and Tafjord]{clark2018think}
Peter Clark, Isaac Cowhey, Oren Etzioni, Tushar Khot, Ashish Sabharwal, Carissa
  Schoenick, and Oyvind Tafjord.
\newblock Think you have solved question answering? try arc, the ai2 reasoning
  challenge.
\newblock \emph{arXiv preprint arXiv:1803.05457}, 2018.

\bibitem[Gebru et~al.(2021)Gebru, Morgenstern, Vecchione, Vaughan, Wallach,
  Iii, and Crawford]{gebru2021datasheets}
Timnit Gebru, Jamie Morgenstern, Briana Vecchione, Jennifer~Wortman Vaughan,
  Hanna Wallach, Hal~Daum{\'e} Iii, and Kate Crawford.
\newblock Datasheets for datasets.
\newblock \emph{Communications of the ACM}, 64\penalty0 (12):\penalty0 86--92,
  2021.

\bibitem[Giunchiglia et~al.(2022)Giunchiglia, Stoian, Khan, Cuzzolin, and
  Lukasiewicz]{giunchiglia2022road}
Eleonora Giunchiglia, Mihaela~C{\u{a}}t{\u{a}}lina Stoian, Salman Khan, Fabio
  Cuzzolin, and Thomas Lukasiewicz.
\newblock Road-r: The autonomous driving dataset with logical requirements.
\newblock \emph{arXiv preprint arXiv:2210.01597}, 2022.

\bibitem[Gr{\"u}nwald(2007)]{grunwald2007minimum}
Peter~D Gr{\"u}nwald.
\newblock \emph{The minimum description length principle}.
\newblock MIT press, 2007.

\bibitem[Guan et~al.(2020)Guan, Jin, Guo, Wang, and
  Cheng]{guan-etal-2020-neuinfer}
Saiping Guan, Xiaolong Jin, Jiafeng Guo, Yuanzhuo Wang, and Xueqi Cheng.
\newblock {N}eu{I}nfer: Knowledge inference on {N}-ary facts.
\newblock In \emph{Proceedings of the 58th Annual Meeting of the Association
  for Computational Linguistics}, pages 6141--6151, Online, July 2020.
  Association for Computational Linguistics.
\newblock \doi{10.18653/v1/2020.acl-main.546}.
\newblock URL \url{https://aclanthology.org/2020.acl-main.546}.

\bibitem[Hamilton et~al.(2018)Hamilton, Bajaj, Zitnik, Jurafsky, and
  Leskovec]{hamilton2018embedding}
Will Hamilton, Payal Bajaj, Marinka Zitnik, Dan Jurafsky, and Jure Leskovec.
\newblock Embedding logical queries on knowledge graphs.
\newblock \emph{Advances in neural information processing systems}, 31, 2018.

\bibitem[Hogan et~al.(2021)Hogan, Blomqvist, Cochez, d’Amato, Melo,
  Gutierrez, Kirrane, Gayo, Navigli, Neumaier, et~al.]{hogan2021knowledge}
Aidan Hogan, Eva Blomqvist, Michael Cochez, Claudia d’Amato, Gerard~De Melo,
  Claudio Gutierrez, Sabrina Kirrane, Jos{\'e} Emilio~Labra Gayo, Roberto
  Navigli, Sebastian Neumaier, et~al.
\newblock Knowledge graphs.
\newblock \emph{ACM Computing Surveys (CSUR)}, 54\penalty0 (4):\penalty0 1--37,
  2021.

\bibitem[Hudson and Manning(2019)]{hudson2019gqa}
Drew~A Hudson and Christopher~D Manning.
\newblock Gqa: A new dataset for real-world visual reasoning and compositional
  question answering.
\newblock In \emph{Proceedings of the IEEE Conference on Computer Vision and
  Pattern Recognition (CVPR)}, pages 6700--6709, 2019.

\bibitem[Johnson et~al.(2017)Johnson, Hariharan, van~der Maaten, Fei-Fei,
  Zitnick, and Girshick]{johnson2017clevr}
Justin Johnson, Bharath Hariharan, Laurens van~der Maaten, Li~Fei-Fei,
  C.~Lawrence Zitnick, and Ross Girshick.
\newblock Clevr: A diagnostic dataset for compositional language and elementary
  visual reasoning.
\newblock In \emph{Proceedings of the IEEE Conference on Computer Vision and
  Pattern Recognition (CVPR)}, 2017.

\bibitem[Kingma and Ba(2014)]{kingma2014adam}
Diederik~P Kingma and Jimmy Ba.
\newblock Adam: A method for stochastic optimization.
\newblock \emph{arXiv preprint arXiv:1412.6980}, 2014.

\bibitem[Krishna et~al.(2017)Krishna, Zhu, Groth, Johnson, Hata, Kravitz, Chen,
  Kalantidis, Li, Shamma, et~al.]{krishna2017visual}
Ranjay Krishna, Yuke Zhu, Oliver Groth, Justin Johnson, Kenji Hata, Joshua
  Kravitz, Stephanie Chen, Yannis Kalantidis, Li-Jia Li, David~A Shamma, et~al.
\newblock Visual genome: Connecting language and vision using crowdsourced
  dense image annotations.
\newblock In \emph{Proceedings of the 30th AAAI Conference on Artificial
  Intelligence}, pages 4088--4095, 2017.

\bibitem[Lake and Baroni(2018)]{lake2018generalization}
Brenden Lake and Marco Baroni.
\newblock Generalization without systematicity: On the compositional skills of
  sequence-to-sequence recurrent networks.
\newblock In \emph{International conference on machine learning}, pages
  2873--2882. PMLR, 2018.

\bibitem[Noy et~al.(2019)Noy, Gao, Jain, Narayanan, Patterson, and
  Taylor]{noy2019industry}
Natasha Noy, Yuqing Gao, Anshu Jain, Anant Narayanan, Alan Patterson, and Jamie
  Taylor.
\newblock Industry-scale knowledge graphs: Lessons and challenges: Five diverse
  technology companies show how it’s done.
\newblock \emph{Queue}, 17\penalty0 (2):\penalty0 48--75, 2019.

\bibitem[Ren and Leskovec(2020)]{ren2020beta}
Hongyu Ren and Jure Leskovec.
\newblock Beta embeddings for multi-hop logical reasoning in knowledge graphs.
\newblock \emph{Advances in Neural Information Processing Systems},
  33:\penalty0 19716--19726, 2020.

\bibitem[Ren et~al.(2020)Ren, Hu, and Leskovec]{ren2020query2box}
Hongyu Ren, Weihua Hu, and Jure Leskovec.
\newblock Query2box: Reasoning over knowledge graphs in vector space using box
  embeddings.
\newblock \emph{arXiv preprint arXiv:2002.05969}, 2020.

\bibitem[Ren et~al.(2023)Ren, Galkin, Cochez, Zhu, and Leskovec]{ren2023neural}
Hongyu Ren, Mikhail Galkin, Michael Cochez, Zhaocheng Zhu, and Jure Leskovec.
\newblock Neural graph reasoning: Complex logical query answering meets graph
  databases.
\newblock \emph{arXiv preprint arXiv:2303.14617}, 2023.

\bibitem[Rissanen(1978)]{rissanen1978modeling}
Jorma Rissanen.
\newblock Modeling by shortest data description.
\newblock \emph{Automatica}, 14\penalty0 (5):\penalty0 465--471, 1978.

\bibitem[Ruffinelli et~al.(2019)Ruffinelli, Broscheit, and
  Gemulla]{ruffinelli2019you}
Daniel Ruffinelli, Samuel Broscheit, and Rainer Gemulla.
\newblock You can teach an old dog new tricks! on training knowledge graph
  embeddings.
\newblock In \emph{International Conference on Learning Representations}, 2019.

\bibitem[Santoro et~al.(2017)Santoro, Raposo, Barrett, Malinowski, Pascanu,
  Battaglia, and Lillicrap]{santoro2017simple}
Adam Santoro, David Raposo, David~GT Barrett, Mateusz Malinowski, Razvan
  Pascanu, Peter Battaglia, and Timothy Lillicrap.
\newblock A simple neural network module for relational reasoning.
\newblock \emph{Advances in Neural Information Processing Systems}, 30, 2017.

\bibitem[Saxton et~al.(2019)Saxton, Grefenstette, Hill, and
  Kohli]{saxton2019analysing}
David Saxton, Edward Grefenstette, Felix Hill, and Pushmeet Kohli.
\newblock Analysing mathematical reasoning abilities of neural models.
\newblock \emph{arXiv preprint arXiv:1904.01557}, 2019.

\bibitem[Suhr and Artzi(2017)]{suhr2017corpus}
Alane Suhr and Yoav Artzi.
\newblock A corpus of natural language for visual reasoning.
\newblock In \emph{Proceedings of the 55th Annual Meeting of the Association
  for Computational Linguistics (ACL)}, volume~1, pages 217--231, 2017.

\bibitem[Trouillon et~al.(2016)Trouillon, Welbl, Riedel, Gaussier, and
  Bouchard]{trouillon2016complex}
Th{\'e}o Trouillon, Johannes Welbl, Sebastian Riedel, {\'E}ric Gaussier, and
  Guillaume Bouchard.
\newblock Complex embeddings for simple link prediction.
\newblock In \emph{International conference on machine learning}, pages
  2071--2080. PMLR, 2016.

\bibitem[Vrande{\v{c}}i{\'c} and Kr{\"o}tzsch(2014)]{vrandevcic2014wikidata}
Denny Vrande{\v{c}}i{\'c} and Markus Kr{\"o}tzsch.
\newblock Wikidata: a free collaborative knowledgebase.
\newblock \emph{Communications of the ACM}, 57\penalty0 (10):\penalty0 78--85,
  2014.

\bibitem[Wang et~al.(2021)Wang, Wang, Lyu, and Zhu]{wang-etal-2021-link}
Quan Wang, Haifeng Wang, Yajuan Lyu, and Yong Zhu.
\newblock Link prediction on n-ary relational facts: A graph-based approach.
\newblock In \emph{Findings of the Association for Computational Linguistics:
  ACL-IJCNLP 2021}, pages 396--407, Online, August 2021. Association for
  Computational Linguistics.
\newblock \doi{10.18653/v1/2021.findings-acl.35}.
\newblock URL \url{https://aclanthology.org/2021.findings-acl.35}.

\bibitem[Wen et~al.(2016)Wen, Li, Mao, Chen, and Zhang]{wen2016representation}
Jianfeng Wen, Jianxin Li, Yongyi Mao, Shini Chen, and Richong Zhang.
\newblock On the representation and embedding of knowledge bases beyond binary
  relations.
\newblock \emph{arXiv preprint arXiv:1604.08642}, 2016.

\bibitem[Weston et~al.(2015)Weston, Bordes, Chopra, Rush, van Merri{\"e}nboer,
  Joulin, and Mikolov]{weston2015towards}
Jason Weston, Antoine Bordes, Sumit Chopra, Alexander~M. Rush, Bart van
  Merri{\"e}nboer, Armand Joulin, and Tomas Mikolov.
\newblock Towards ai-complete question answering: A set of prerequisite toy
  tasks.
\newblock \emph{arXiv preprint arXiv:1502.05698}, 2015.

\bibitem[Yang et~al.(2014)Yang, Yih, He, Gao, and Deng]{yang2014embedding}
Bishan Yang, Wen-tau Yih, Xiaodong He, Jianfeng Gao, and Li~Deng.
\newblock Embedding entities and relations for learning and inference in
  knowledge bases.
\newblock \emph{arXiv preprint arXiv:1412.6575}, 2014.

\bibitem[Yang et~al.(2018)Yang, Li, and Zhu]{yang2018dataset}
Jiale Yang, Jinnan Li, and Yuke Zhu.
\newblock A dataset and architecture for visual reasoning with a working
  memory.
\newblock \emph{arXiv preprint arXiv:1803.06092}, 2018.

\bibitem[Zhang et~al.(2018)Zhang, Zhang, and Lapata]{zhang2018metaqa}
Xianda Zhang, Yifan Zhang, and Mirella Lapata.
\newblock Metaqa: A dataset of metaphorically annotated movieqa questions.
\newblock In \emph{Proceedings of the 27th International Conference on
  Computational Linguistics}, pages 1954--1964, 2018.

\end{thebibliography}

\newpage

\section{Supplementary Material}

\localtableofcontents

\subsection{Datasets for Neurosymbolic Methods}

Neurosymbolic methods aim to combine neural networks with symbolic representations. As mentioned in Section \ref{related-work-section}, several datasets already exist in the literature for evaluating the performance of neurosymbolic methods. Table \ref{tab:comparison} highlights widely used datasets used for benchmarking neurosymbolic systems.

\nopagebreak

\begin{table}[ht]
\tiny
\centering
\caption{Brief comparison of commonly used datasets for benchmarking neurosymbolic methods, listed in ascending order of publication year. For each dataset, we provide an overview of the task, domain, modality, key characteristics, and whether the dataset is synthetic.}
\label{tab:comparison}
\begin{tabularx}{\linewidth}{ >{\raggedright\arraybackslash}p{1.5cm} >{\raggedright\arraybackslash}X >{\raggedright\arraybackslash}X>{\raggedright\arraybackslash}X>{\raggedright\arraybackslash}p{2.5cm}>{\raggedright\arraybackslash}p{1.0cm}}
\hline 
\\
{\small Dataset} & {\small Task} & {\small Domain} & {\small Modality} & {\small Key Characteristics} & {\small Synthetic}
\\ \\
\hline
\textbf{bAbI} \\ \cite{weston2015towards} & Language Reasoning & Natural Language & Text & Basic reasoning, generalization & Yes \\ \\
\textbf{SNLI} \\ \cite{bowman2015large} & Logical Reasoning & Natural Language & Text & Entailment, contradiction, neutral relationships & No \\ \\
\textbf{CLEVR} \\ \cite{johnson2017clevr} & Visual Reasoning & Computer Vision & Images \& Text & Object counting, comparison, querying attributes & Yes \\ \\
NLVR \\ \cite{suhr2017corpus} & Visual Reasoning & Computer Vision & Images \& Text & Visual reasoning, natural language understanding & Yes \\ \\
\textbf{Sort-of-CLEVR} \\ \cite{santoro2017simple} & Relational Reasoning & Computer Vision & Images \& Text & Spatial and relational reasoning & Yes \\ \\
Visual Genome \\ \cite{krishna2017visual} & Visual Reasoning & Computer Vision & Images \& Text & Object recognition, relationships, attributes & No \\ \\
\textbf{Aristo} \\ \cite{clark2018think} & Science Reasoning & Natural Language & Text & Natural language understanding, applying knowledge & No \\ \\
\textbf{COG} \\ \cite{yang2018dataset} & Cognitive Capabilities & Computer Vision & Images \& Text & Temporal and logical reasoning & Yes \\ \\
\textbf{MetaQA} \\ \cite{zhang2018metaqa} & Multi-hop Reasoning & Graph & Knowledge Graph & Multi-step reasoning, knowledge base & No \\ \\
\textbf{SCAN} \\ \cite{lake2018generalization} & Compositional Generalization & Command-based Language & Text & Understanding and generating novel commands & No \\ \\
\textbf{Math Dataset} \\ \cite{saxton2019analysing} & Math Reasoning & Natural Language & Text & Language understanding, symbolic reasoning & No \\ \\
\textbf{GQA} \\ \cite{hudson2019gqa} & Visual Reasoning & Computer Vision & Images & Text \& Spatial and relational reasoning & No \\ \\
\textbf{ROAD-R} \\ \cite{giunchiglia2022road} & Visual Reasoning & Computer Vision & Videos \& (handcrafted) Logical Rules & Logical reasoning & No \\
\hline
\end{tabularx}
\end{table} 

\subsection{Reproducibility Statement}
To make our work fully reproducible, we make the codebase of our experiments public and open. Our code is available on \url{https://github.com/thiviyanT/IntelliGraphs}. For each experiment, we also provide the hyperparameter configurations we used. Furthermore, we have released a new Python package for interacting with the IntelliGraphs datasets through the following software package repositories: \textbf{conda} (\url{https://anaconda.org/thiv/intelligraphs}) and \textbf{pypi} (\url{https://pypi.org/project/intelligraphs}). To ensure long-term preservation and easy access, we made the datasets available on Zenodo (\url{https://doi.org/10.5281/zenodo.7824818}). Experimental details can be found in the next Section.

\subsection{Experimental Details}
We used the PyTorch library \footnote{\url{https://pytorch.org/}} to develop and test the models. All experiments were performed on a single-node machine with an Intel(R) Xeon(R) Gold 5118 (2.30GHz, 12 cores) CPU and 64GB of RAM, with four NVIDIA RTX A4000 GPUs (16GB of VRAM). We used PyTorch's GPU acceleration for training the models. We used the Adam optimiser with variable learning rates \citep{kingma2014adam}.

\subsubsection{Hyperparameters}
\label{subsec:benchmarks}

For each dataset, we performed hyperparameter sweeps using every baseline model (TransE, DistMult, ComplEx) using Weights\&Biases \footnote{\url{https://wandb.ai/}}. For this, we used a random search strategy with the goal of finding the hyperparameter configurations that yield the minimum compression bits on the validation set. We do not include the reciprocal relation model, and we used the highest batch size that we could fit in memory. Table \ref{tab:hyperparameter} shows the hyperparameter values we obtained via the sweeps. The random baseline did not require hyperparameter finetuning. We also used Weights \& Biases for monitoring our experiments. 

\begin{table}[ht]
\centering
\caption{The results of a random hyperparameter search, presenting the chosen hyperparameters for different datasets and baseline models. The hyperparameters include \emph{batch size, embedding size, learning rate, biases usage, and initialization method}. The batch size indicates the number of training subgraphs processed together before updating the model. The embedding size represents the dimensionality of the entity and relation embeddings. The learning rate controls the step size taken during model optimization. The biases denote whether bias terms are included in the model, and the initialization method refers to the technique used to initialize the model's parameters.}
\begin{tabular}{l l l l l l l}
\hline
\textbf{Dataset} & \textbf{Model} & \textbf{Batch Size} & \textbf{Emb.} & \textbf{Learning Rate} & \textbf{Biases} & \textbf{Init.} \\
\hline
syn-paths & transe & 4096 & 1531 & 7.029817939842623e-05 & \texttt{False} & \texttt{uniform} \\
syn-paths & distmult & 4096 & 158 & 0.0697979730927795 & \texttt{False} & \texttt{uniform} \\
syn-paths & complex & 4096 & 587 & 5.264944612887405e-05 & \texttt{False} & \texttt{uniform} \\
\hline
syn-tipr & transe & 2048 & 147 & 0.0008716274682049251 & \texttt{True} & \texttt{normal} \\
syn-tipr & distmult & 2048 & 168 & 0.005497983171450242 & \texttt{True} & \texttt{normal} \\
syn-tipr & complex & 2048 & 350 & 0.0015597556675205502 & \texttt{True} & \texttt{normal} \\
\hline
syn-types & transe & 2048 & 376 & 0.003017403610019781 & \texttt{True} & \texttt{uniform} \\
syn-types & distmult & 2048 & 273 & 0.0006013105272716594 & \texttt{True} & \texttt{uniform} \\
syn-types & complex & 2048 & 996 & 5.603405855158606e-05 & \texttt{False} & \texttt{uniform} \\
\hline
wd-movies & transe & 4096 & 68 & 0.000638003263107625 & \texttt{False} & \texttt{normal} \\
wd-movies & distmult & 4096 & 181 & 0.00307853821840767 & \texttt{True} & \texttt{uniform} \\
wd-movies & complex & 4096 & 102 & 0.019520125878695407 & \texttt{False} & \texttt{uniform} \\
\hline
wd-articles & transe & 32 & 888 & 6.094053758340765e-05 & \texttt{True} & \texttt{normal} \\
wd-articles & distmult & 32 & 65 & 0.03833121378755901 & \texttt{False} & \texttt{uniform} \\
wd-articles & complex & 32 & 283 & 0.002251396972378282 & \texttt{False} & \texttt{normal} \\
\hline
\end{tabular}
\label{tab:hyperparameter}
\end{table}

\subsection{Semantics of IntelliGraphs}
\label{appx-semantics}

Logical rules provide a formal framework for expressing and reasoning about the semantics of a system. In this section, we discuss the logical rules we use to verify the semantics of the IntelliGraphs datasets. We express each logical rule using First-Order Logic (FOL) unless otherwise stated. We opted for First Order Logic (FOL) as the formal language to communicate logical constraints due to its ability to effectively express the necessary constraints and its widespread understanding within the machine learning community \footnote{These FOL logical constraints can also be rewritten into data specification languages, such as DataLog.}. 

Although we provide the general FOL rules to check the semantics of graphs of \emph{any arbitrary lengths}, we apply a size constraint (\emph{i.e.} checking for graphs with a fixed number of triples) for the synthetic datasets. This is because the synthetic data generator produces graphs with fixed length and we defined it as part of our semantics. The size constraint can also be expressed in FOL, but we specify this constraint in \emph{natural language} for brevity.

Traditionally, a reasoning engine is used to check logical consistencies in knowledge bases. 
We wrote a semantic checker in Python. This was more convenient to use within our framework as the graphs could be evaluated, without having to manually load them into a reasoning engine individually. Our semantic checker was written to closely follow the logical rules, and it is accessible through the IntelliGraph python package. 

\subsubsection{Logical Rules of \texttt{syn-paths}}

\nopagebreak


\begin{align*}
\forall x,y,z &: connected(x,y) \land connected(y,z) \Rightarrow connected(x,z) \\ 
\forall x,y &: edge(x,y) \Rightarrow connected(x,y) \\ 
\exists x &: root(x) \\ 
\forall x,y &: root(x) \land root(b) \Rightarrow a=b \\ 
\forall x &: root(x) \Leftrightarrow \forall y: \neg  edge(y,x) \\ 
\forall x,y &: connected(x,y) \Rightarrow x \not= y \\ 
\forall x &: root(x) \Rightarrow \forall y: ( connected(x,y) \lor x=y) \\ 
\forall x,y,z &: edge(y,x) \land edge(z,x) \Rightarrow y=z \\ 
\forall x,y,z &: edge(x,y) \land edge(x,z) \Rightarrow y=z \\ 
\forall x,y &: edge(x,y) \Leftrightarrow cycle\_to(x,y) \lor drive\_to(x,y) \lor train\_to(x,y) \\
\text{\emph{Number of edges}} &: 3
\end{align*}

\newpage

\subsubsection{Logical Rules of \texttt{syn-types}}

\nopagebreak


\begin{align*}
\forall x,y &: spoken\_in(x, y) \Rightarrow language(x) \land country(y)  \\ 
\forall x,y &: could\_be\_part\_of(x, y) \Rightarrow city(x) \land country(y))  \\ 
\forall x,y &: (same\_type\_as(x, y) \Rightarrow (language(x)\land language(y)) \\
&\quad\quad\lor (city(x) \land city(y)) \lor (country(x) \land country(y))  \\ 
\forall x &: language(x) \Rightarrow \neg country(x) \land \neg city(x)  \\ 
\forall x &: country(x) \Rightarrow \neg language(x) \land \neg city(x)  \\ 
\forall x &: city(x) \Rightarrow \neg language(x) \land \neg country(x)  \\ 
\text{\emph{Number of edges}} &: 3
\end{align*}










\subsubsection{Logical Rules of \texttt{syn-tipr}}

\nopagebreak

%

\begin{align*}
\forall x,y &: has\_role(x, y) \Rightarrow academic(x) \land role(y) \\
\forall x,y &: has\_name(x, y) \Rightarrow academic(x) \land name(y) \\
\forall x,y &: has\_time(x, y) \Rightarrow academic(x) \land time(y) \\
\forall x,y &: start\_year(x, y) \Rightarrow time(x) \land year(y) \\
\forall x,y &: end\_year(x, y) \Rightarrow time(x) \land year(y) \\
\forall x,y,z &: end\_year(x, y) \land start\_year(x, z) \Rightarrow before(y, z) \\
\forall x &: \neg has\_role(x, x) \\
\forall x &: \neg has\_name(x, x) \\
\forall x &: \neg has\_time(x, x) \\
\forall x &: \neg start\_year(x, x) \\
\forall x &: \neg end\_year(x, x) \\
\forall x &: academic(x) \Rightarrow \neg role(x) \land \neg time(x) \land \neg name (x) \land \neg year(x) \\
\forall x &: role(x) \Rightarrow \neg academic(x) \land \neg time(x) \land \neg name (x) \land \neg year(x) \\
\forall x &: time(x) \Rightarrow \neg academic(x) \land \neg role(x) \land \neg name (x) \land \neg year(x) \\
\forall x &: year(x) \Rightarrow \neg academic(x) \land \neg role(x) \land \neg name (x) \land \neg time(x) \\
\forall x &: name(x) \Rightarrow \neg academic(x) \land \neg role(x) \land \neg year (x) \land \neg time(x) \\
\text{\emph{Number of edges}} &: 5
\end{align*}

\subsubsection{Logical Rules of \texttt{wd-movies}}

\nopagebreak

\begin{align*}
\forall x,y &: connected(x,y) \Leftrightarrow has\_director(x,y) \lor has\_actor(x,y) \lor has\_genre(x,y) \\
\exists x &: has\_director(x, \texttt{existential\_node}) \\
\exists x &: has\_actor(x, \texttt{existential\_node}) \\
\exists x &: has\_genre(x, \texttt{existential\_node}) \\ 
\forall x &: x \neq \texttt{existential\_node} \Rightarrow connected(\texttt{existential\_node}, x) \\
\forall x,y &: x \neq \texttt{existential\_node} \land y \neq \texttt{existential\_node} \Rightarrow \neg connected(x, y) \\
\forall x &: \neg connected(x, \texttt{existential\_node}) \\
\forall x,y &: has\_director(x,y) \lor has\_actor(x,y) \Rightarrow person(y) \\
\forall x &: \neg person(x) \lor \neg genre(x) \\
\forall x,y &: has\_genre(x,y) \Rightarrow genre(y) 
\end{align*}

\subsubsection{Logical Rules of \texttt{wd-articles}}

\nopagebreak

\begin{align*}
\exists x &: has\_author(\texttt{article\_node}, x) \\
\forall x,y &: connected(x,y) \Leftrightarrow has\_author(x,y) \lor has\_name(x,y) \lor has\_order(x,y) \lor \\
& \quad cites(x,y) \lor has\_subject(x,y) \lor subclass\_of(x,y)  \\
\forall x,y &: connected(x,y) \Rightarrow \neg connected(y,x) \lor cites(y, x)  \\
\forall x &: \neg connected(x, x)  \\
\forall x, y &: has\_author(x, y) \Rightarrow x = \texttt{article\_node}  \\
& \quad article(\texttt{article\_node}) \lor iri(\texttt{article\_node})  \\
\forall x &: has\_author(\texttt{article\_node}, x) \Rightarrow authorpos(x)  \\
\forall x &: authorpos(x) \Leftrightarrow \exists y: has\_order(x, y) \land \exists y: has\_name(x, y)  \\
\forall x,y &: has\_order(x,y) \Rightarrow authorpos(x) \land ordinal(y)  \\
\forall x, y &: has\_name(x, y) \Rightarrow authorpos(x) \land name(y) \lor iri(y)  \\
\forall x, y, z &: has\_order(x, y) \land has\_order(x, z) \Rightarrow y = z  \\
\forall x, y, z &: has\_name(x, y) \land has\_order(x, z) \Rightarrow y = z  \\
\forall x &: author(x) \Rightarrow \neg subject(x) \land \neg iri(x) \land \neg name(x) \land \neg ordinal(x) \land \neg author\_pos(x)  \\
\forall x &: subject(x) \Rightarrow \neg author(x) \land \neg iri(x) \land \neg name(x) \land \neg ordinal(x) \land \neg author\_pos(x)  \\
\forall x &: iri(x) \Rightarrow \neg author(x) \land \neg subject(x) \land \neg name(x) \land \neg ordinal(x) \land \neg author\_pos(x)  \\
\forall x &: name(x) \Rightarrow \neg subject(x) \land \neg iri(x) \land \neg author(x) \land \neg ordinal(x) \land \neg author\_pos(x)  \\
\forall x &: ordinal(x) \Rightarrow \neg subject(x) \land \neg iri(x) \land \neg name(x) \land \neg author(x) \land \neg author\_pos(x)  \\
\forall x &: author\_pos(x) \Rightarrow \neg subject(x) \land \neg iri(x) \land \neg name(x) \land \neg author(x) \land \neg ordinal(x)  \\
\forall x,y,z &: subclass\_trans(x, y) \land subclass\_trans(y, z) \Rightarrow subclass\_trans(x, z)  \\
\forall x,y &: subclass\_of(x,y) \Rightarrow subclass\_trans(x,y) \land \\
&\phantom{: subclass\_of(x,y) \Rightarrow }(iri(x) \lor subject(x)) \land (iri(y) \lor subject(y))  \\
\forall x,y &: subclass\_of(x,y) \Rightarrow \exists z: subclass\_trans(x,z) \land has\_subject(\texttt{article\_node}, z)  \\
\forall x,y &: cites(x, y) \Rightarrow iri(y) \land x=\texttt{article\_node}  \\
\forall x,y &: has\_subject(x, y) \Rightarrow (subject(y) \lor iri(y)) \land x=\texttt{article\_node}
\end{align*}

\nopagebreak

In addition to the aforementioned rules for \texttt{wd-articles}, our semantic checker checks the ordinal of the author's position to make sure that they are a complete list of  consecutive numbers (\emph{i.e.} \texttt{ordinal\_000}, \texttt{ordinal\_001}, \texttt{ordinal\_002}, ..., etc.), but we leave it out of the rules for brevity.

\subsection{Synthetic Dataset Generation}

The synthetic dataset generator contains two main modules: 1) a \emph{triple sampler} is a module that samples new triples one by one, 2) a \emph{triple verifier} module checks each triple for semantic validity before they are added to a subgraph. The generator builds a subgraph by sampling one triple at a time and verifiying. If the triple passes the semantic check, it is added to a subgraph. To avoid duplicate triples within the sam subgraph, we check if triple already exists before adding it to a subgraph. This is done until a certain number of valid triples are samples. For reproducibility, we use the same seed for all random data generations ($\verb|seed=42|$).For each dataset, we generate \emph{training}, \emph{validation} and \emph{test} sets. To avoid data leakage, we check that these graphs are unique before splitting the dataset. In this section, we briefly describe how IntelliGraphs efficiently samples valid subgraphs.

\subsubsection{\texttt{syn-paths}}
The entities are labelled after 49 Dutch cities and the relations are different modes of transport ($\verb|train_to, drive_to, cycle_to|$). This dataset primarily checks whether baseline models can do structure learning.

A path graph $P_{k}(G)$ of a graph $G$ has vertex set $\Pi_{k}(G)$ and edges joining pairs of vertices that represent two paths $P_{k}$, the union of which forms either a path $P_{k+1}$. We denote by $\Pi_{k}(G)$ the set of all paths of $G$ on $k$ vertices $(k \geq 1)$, and we randomly sample $n$ edges from $\Pi_{k}(G)$ to generate each path graph.

To generate a path graph, we begin by selecting a head (\emph{i.e.} source node) by randomly selecting a Dutch city and then we sample relation and a tail (\emph{i.e.} target node). For the next triple in the subgraph, we use the previous target node as the source node and then sample a relation and a target node. We can repeat the last step $k-2$ number of times to build a path-graph with $k$ edges. We ensure that each subgraph includes all three different relations. We avoid generating cyclic path-graphs. 

\subsubsection{\texttt{syn-types}}
This dataset contains three types of entities ($\verb|cities|$, $\verb|countries|$ \& $\verb|languages|$), 30 entities in total (10 instances of each entity type), and three relations ($\verb|same_type_as|$, $\verb|could_be_part_of|$ \& $\verb|could_be_spoken_in|$). This dataset primarily checks whether baseline models can learn the types of entities correctly.

For each relation, we sample a head and a tail entity of the corresponding type. For instance, for the relation $\verb|could_be_spoken_in|$ we sample a language for the head of a triple and a country for the tail. Similarly, we sample other triples to be added to the same subgraph, until a certain number of valid triples have been sampled.

It is important to note that the \texttt{syn-types} dataset is not meant to be factually accurate but rather serves as a way to study the type semantics learned by machine learning models.

\subsubsection{\texttt{syn-tipr}}

This datasets contain three entity types (\texttt{names}, \texttt{roles}, \texttt{years}) and (\texttt{has\_role}, \texttt{has\_name}, \texttt{start\_year} and \texttt{end\_year}). We used a random name generator \footnotetext{\url{https://www.behindthename.com/random/}} to generate 50 names. For simplicity, we treat \emph{years} are entities rather than literals. In each subgraph, there are two existential nodes: $\verb|_academic|$ and $\verb|_time|$). The main purpose of this dataset is to check structure learning and check \emph{basic} temporal reasoning (in this case, whether \texttt{end\_year} appears after \texttt{start\_year}). 

The subgraphs in this dataset was modeeled after the \emph{time-indexed person role} (tipr) pattern in Semantic Web. For generating these subgraphs, we take the tipr pattern as a template and randomly sampled entities the correct entity type. For instance, the relation \texttt{has\_role} would always have an academic\_node in the head position of a triple and a role as a tail. Similarly, we sample triples for the other relations (\texttt{has\_name}, \texttt{has\_time}, \texttt{start\_year}, \texttt{end\_year}). Valid triples containing every relations is sampled. In total, every subgraph will contain five triples. 

\subsection{Wikidata Dataset Generation}

For reproducibility, we use a specific Wikidata dump to extract the data, rather than the live version. For both datasets, we use the Wikidata HDT dump from 3 March 2021, available from the HDT website \footnotemark.

\footnotetext{\url{https://www.rdfhdt.org/datasets/}}

In both cases, we first extract all data that fits the template of the graph, for instance, for every movie we extract all actors, directors and genres. We then \emph{prune} this data to ensure that every entity occurs in enough instances to allow a model to learn a representation for it. Depending on the dataset we either remove the infrequent nodes or replace them by existential nodes. We set the minimal frequency to 6 in both datasets.

To avoid the situation where certain entity nodes are only present in the validation or test data, we must make our splits carefully. Ideally, we'd like for each entity to be present in all three splits of the data, and where this is not possible, for it to be present in at least the training data.

To achieve this, we use the following algorithm: for each instance we collect ``votes'' among all its entities for which of the three splits it should be part of. Simultaneously, for each entity, we collect the splits of which it is a member. The aim is to have all entity in each instance vote for the same split, and for each entity to be represented in all splits. We first alternate fixing one of the two problems: we unify the votes by choosing a random entity and setting the votes of the other entities in the instance to that vote. After all votes have been fixed, we fix the split memberships by, for each entity that is not representing in all splits, taking the most frequent split and changing the vote of one of its instances to the missing split, repeating until all splits are represented.

We alternate these steps for 50 iterations. Then, in the first step, we move any instance with conflicting votes to the training data and repeat the iteration in this fashion for another 20 steps. For both datasets, this leads to all entities being represented in the training data, and only a small number present in only the test or only the training data.

For both datasets, the labels are Wikipedia IRIs, but a mapping to human-readable labels is provided. In this paper, we replace IRIs with these for readability.

\subsubsection{\texttt{wd-movies}}

We collect all entities that are labeled as ``instance of'' the class ``film''. For each we extract all entities connected by the relations ``cast member'', ``director'' and ``genre'' as its actors, directors and genres respectively.

We then prune the data by removing all actors, directors and genres that do not appear in at least 6 instances. We then remove any movies that are left with no actors or no directors. We allow movies with no genres. We iterate these two steps until no changes are made. Finally, we make a test, train and validation split by the process described above. The following Wikidata properties and entities were used: 

\begin{tabular}{l l}
\hline
\textbf{label} & \textbf{wikidata IRI} \\
\hline
instance of & \url{http://www.wikidata.org/prop/P31} \\
film & \url{http://www.wikidata.org/entity/Q11424} \\
cast member & \url{http://www.wikidata.org/prop/P161} \\
director & \url{http://www.wikidata.org/prop/P57} \\
genre & \url{http://www.wikidata.org/prop/P136} \\
\hline
\end{tabular}

\subsubsection{\texttt{wd-articles}}

We collect all entities from wikidata that are the object of a triple with the relation ``cites''. 

For each article we collect the full list of authors, using the relations ``author'' and ``author name string''. The former is used to refer to authors that are represented in Wikidata as an entity, and the latter is used for authors represented only by their name as a string literal. We require at least one of the authors to be represented by an entity. If not, the article is filtered out.

Such statements are commonly annotated in Wikidata with an \emph{ordinal}, representing the order of the author in the author list. We extract these as well. If any author does not have an ordinal or if the collection of these ordinals does not coincide exactly with the sequence $1, \ldots, n$, with $n$ the number of authors, the article is filtered out.

We then collect all articles that, as recorded in Wikidata, the current article cites. If there are no such references, the article is filtered out.

Finally, we collect the article subjects, and for each subject, every superclass and its superclass, that are an instance of ``academic discipline''. We do not filter based on the subjects (no subjects or superclasses is allowed).

We collect the first 100 000 such articles for the dataset \texttt{wd-articles}, and all such articles for the dataset \texttt{wd-articles-large}.

As with \texttt{wd-movies}, we prune the data to eliminate any entities that occur in fewer than 6 instances. For the authors, the article itself and the subjects, we replace these with existential nodes. These have node labels specific to the role they play in the graph: \texttt{\_article}, \texttt{\_author001}, and \texttt{\_subject001}. Any references to infrequent entities are removed. As before. this removal process is iterated until the dataset stabilizes.

Splits are then made using the algorithm described above. In the construction of the dataset, we add authors by introducing a blank node (using label \texttt{\_authorpos} and the relation \texttt{has\_author}), to which the author identity (\texttt{has\_name}) and the ordinal (\texttt{has\_order}) are connected. References are added by a single edge with the relation \texttt{cites} and subjects and superclasses with the relations \texttt{has\_subject} and \texttt{subclass\_of}.

The following Wikidata properties were used.

\nopagebreak

\begin{tabular}{l l}
\hline
\textbf{label} & \textbf{wikidata IRI} \\
\hline
cites & \url{http://www.wikidata.org/prop/P2860} \\
author & \url{http://www.wikidata.org/prop/P50} \\
author name string & \url{http://www.wikidata.org/prop/P2093} \\
main subject & \url{http://www.wikidata.org/prop/P921} \\
subclass of & \url{http://www.wikidata.org/prop/P279} \\
academic discipline & \url{http://www.wikidata.org/entity/Q11862829} \\
\hline
\end{tabular}

\subsubsection{Example subgraphs}

Figure \ref{fig:example-triples} showcases a selection of example subgraph from each dataset: \texttt{syn-paths}, \texttt{syn-tipr}, \texttt{syn-types}, \texttt{wd-movies}, and \texttt{wd-articles}. 

\nopagebreak

\begin{figure*}
\tiny
\centering
\begin{tabularx}{\linewidth}{X}
\toprule
\textbf{syn-paths} \\
\texttt{[Nieuwegein drive\_to Lelystad, Lelystad drive\_to IJmuiden, IJmuiden cycle\_to Zaanstad]} \\
\texttt{[IJmuiden cycle\_to Maastricht,  Maastricht train\_to Roermond, Roermond drive\_to Groningen]} \\
\texttt{[Hilversum cycle\_to Emmen, Emmen drive\_to Spijkenisse, Spijkenisse train\_to Sittard]} \\
\midrule
\textbf{syn-tipr} \\
\texttt{[\_academic has\_name Cleophas Erős, \_academic has\_role masters researcher, \_academic has\_time \_time, \_time start\_year 2016, \_time end\_year 2018]} \\
\texttt{[\_academic has\_name Romana Sitk, \_academic has\_role professor, \_academic has\_time \_time, \_time start\_year 1982, \_time end\_year 2009]} \\
\texttt{[\_academic has\_name Drusus Krejči, \_academic has\_role assistant professor, \_academic has\_time \_time, \_time start\_year 1996, \_time end\_year 2000]} \\
\texttt{[\_academic has\_name Božidar Bullard, \_academic has\_role professor\_academic has\_time \_time, \_time start\_year 1973, \_time end\_year 1988]} \\
\midrule
\textbf{syn-types} \\
\texttt{[Dutch same\_type\_as English, Budapest could\_be\_part\_of United Kingdom, Czech spoken\_in Serbia]} \\
\texttt{[Serbia same\_type\_as Spain, Paris could\_be\_part\_of Norway, Dutch spoken\_in Greece]} \\
\texttt{[Greek same\_type\_as Italian, Budapest could\_be\_part\_of Ireland, French spoken\_in Serbia]} \\
\midrule
\textbf{wd-movies} \\
\texttt{[\_movie has\_director P. Pullaiah, \_movie has\_actor Gummadi Venkateswara Rao, \_movie has\_actor Akkineni Nageswara Rao, \_movie has\_actor Anjali Devi, \_movie has\_actor Chittoor Nagaiah, \_movie has\_actor Ramana Reddy, \_movie has\_actor Relangi Venkata Ramaiah, \_movie has\_actor S. V. Ranga Rao, \_movie has\_actor Santha Kumari, \_movie has\_genre historical film, \_movie has\_genre biographical film]} \\
\texttt{[\_movie has\_director Albert Brooks, \_movie has\_actor Kathryn Harrold, \_movie has\_actor Albert Brooks, \_movie has\_actor Bruno Kirby, \_movie has\_genre comedy film]} \\
\texttt{[\_movie has\_director Dragoslav Lazić, \_movie has\_actor Vesna Malohodžić, \_movie has\_actor Snežana Savić, \_movie has\_genre comedy film]} \\
\texttt{[\_movie has\_director Balu Mahendra, \_movie has\_actor Silk Smitha, \_movie has\_actor Sridevi, \_movie has\_actor Kamal Haasan, \_movie has\_genre romance film]} \\
\midrule
\textbf{wd-articles} \\
\texttt{ [\_article has\_author \_authorpos000, \_authorpos000 has\_name \_author000, \_authorpos000 has\_order ordinal\_001, \_article has\_author \_authorpos001, \_authorpos001 has\_name \_author001, \_authorpos001 has\_order ordinal\_002, \_article has\_author \_authorpos002, \_authorpos002 has\_name \_author002, \_authorpos002 has\_order ordinal\_003, \_article has\_author \_authorpos003, \_authorpos003 has\_name \_author003, \_authorpos003 has\_order ordinal\_004, \_article has\_author \_authorpos004, \_authorpos004 has\_name \_author004, \_authorpos004 has\_order ordinal\_005, \_article has\_author \_authorpos005, \_authorpos005 has\_name \_author005, \_authorpos005 has\_order ordinal\_006, \_article has\_author \_authorpos006, \_authorpos006 has\_name \_author006, \_authorpos006 has\_order ordinal\_007, \_article has\_author \_authorpos007, \_authorpos007 has\_name \_author007, \_authorpos007 has\_order ordinal\_008, \_article cites http://www.wikidata.org/entity/Q25938995, \_article cites http://www.wikidata.org/entity/Q28242060, \_article cites http://www.wikidata.org/entity/Q28286732, \_article cites http://www.wikidata.org/entity/Q34453213, \_article cites http://www.wikidata.org/entity/Q34541710, \_article cites http://www.wikidata.org/entity/Q35758845, \_article cites http://www.wikidata.org/entity/Q37942996, \_article cites http://www.wikidata.org/entity/Q37972005, \_article cites http://www.wikidata.org/entity/Q42642132, \_article has\_subject http://www.wikidata.org/entity/Q214781, http://www.wikidata.org/entity/Q214781 subclass\_of http://www.wikidata.org/entity/Q413]
    } \\
\bottomrule
\end{tabularx}
\caption{IntelliGraphs contains five datasets: \texttt{syn-paths}, \texttt{syn-tipr}, \texttt{syn-types}, \texttt{wd-movies}, and \texttt{wd-articles}. Here we showcase a few example subgraphs from each dataset. The subgraphs are presented as a list of triples, where each list item represents a subgraph.}
\label{fig:example-triples}
\end{figure*}

\section{Datacard}

An up-to-date version of the data card can be found on \url{https://github.com/thiviyanT/IntelliGraphs/blob/main/Datacard.md}.

\end{document}